\def\year{2020}\relax
\documentclass[letterpaper]{article}
\usepackage{aaai}
\usepackage{times}
\usepackage{helvet}
\usepackage{courier}
\usepackage[utf8]{inputenc} % allow utf-8 input
\usepackage[T1]{fontenc}    % use 8-bit T1 fonts
\usepackage{hyperref}       % hyperlinks
\usepackage{url}            % simple URL typesetting
\usepackage{booktabs}       % professional-quality tables
\usepackage{amsfonts}       % blackboard math symbols
\usepackage{nicefrac}       % compact symbols for 1/2, etc.
\usepackage{microtype}      % microtypography

\usepackage{graphicx, subfigure}
\usepackage{amsmath}
\usepackage{multirow}
\usepackage{amstext}
\usepackage{color}
\usepackage{enumitem}

\graphicspath{ {fig/} }
\begin{document}
\title{Multiple-objective Reinforcement Learning for Inverse Design and Identification}
% \author{submission \# 218}
\author{
Haoran Wei,\textsuperscript{1}
Mariefel Olarte,\textsuperscript{2},
Garrett B. Goh*\textsuperscript{2}\\
\textsuperscript{1}{University of Delaware}\\
\textsuperscript{2}{Pacific Northwest National Laboratory}\\
nancywhr@udel.edu,
mariefel.olarte@pnnl.gov,
garrett.goh@pnnl.gov}
%   \And
%   Mariefel Olarte \\
%   PNNL \\
%   \texttt{mariefel.olarte@pnnl.gov}
%   \And
%   Garrett B. Goh* \\
%   PNNL \\
%   \texttt{garrett.goh@pnnl.gov}
% }
% \author{Haoran Wei,\textsuperscript{1*}
% Yuanbo Wang,\textsuperscript{2*}, Lidia Mangu,\textsuperscript{2**}, Keith Decker,\textsuperscript{1**}\\
% \textsuperscript{1}{University of Delaware}\\
% \textsuperscript{2}{J.P. Morgan}\\
% nancywhr@udel.edu,
% yuanbo.wang@jpmchase.com, 
% lidia.l.mangu@jpmchase.com
% decker@udel.edu}

\maketitle
\begin{abstract}
\begin{quote}
The aim of  inverse chemical design is to develop new molecules with given optimized molecular properties or objectives. Recently, generative deep learning (DL) networks are considered as the state-of-the-art in inverse chemical design and have achieved early success in generating molecular structures with desired properties in the pharmaceutical and material chemistry fields. However, satisfying a large number (> 10 objectives) of molecular objectives is a limitation of current generative models. To improve the model's ability to handle a large number of molecule design objectives, we developed a Reinforcement Learning (RL) based generative framework to optimize chemical molecule generation. Our use of Curriculum Learning (CL) to fine-tune the pre-trained generative network allowed the model to satisfy up to 21 objectives and increase the generative network's robustness. The experiments show that the proposed multiple-objective RL-based generative model can correctly identify unknown molecules with an 83\% $\sim$ 100\% success rate, compared to the baseline approach of  0\%. Additionally, this proposed generative model is not limited to just chemistry research challenges; we anticipate that problems that utilize RL with multiple-objectives will benefit from this framework.

\end{quote}
\end{abstract}

\section{Introduction}

% Traditional chemical design for a target use case, such as a new drug compound, is dependent on using high-throughput experiments or simulations to quickly screen through a list of candidate chemicals, with the hope that some of them will meet the desired design requirements. 
Designing a chemical for specific target use, such as a new drug compound, traditionally relies on high throughput screening experiments or simulations. Given that the chemical space spans on the order of $10^{60} \sim 10^{100}$ potential targets, it comes with no surprise that this brute-force approach is highly iterative with low success rates. Recent advances in deep learning (DL) have demonstrated preliminary success with the inverse design paradigm, where desired properties are used as input, for the DL models to generate chemical structures that would satisfy the design requirements. To date, various generative approaches have been proposed and deployed in the chemical design domain, including models based on variational autoencoders\cite{kingma2013auto}, generative RNN models\cite{olivecrona2017molecular} and GANs \cite{guimaraes2017objective}.

Besides chemical design, generative models can also be used for chemical identification, which has various applications in fields such as in the production of biofuels, where complex mixtures of compounds from biomass are generated. Current approaches used in identifying chemicals are limited, often relying on database matching of Mass Spectrometry (MS) and Nuclear Magnetic Resonance (NMR) spectra to known chemicals \cite{bingol2018recent,boiteau2018structure}. However, given the extensiveness of the chemical space, this approach would be effective only in identifying the chemicals already available in the database. On the other hand, optimization of a target compound structure against a set of constraints (i.e., fingerprints), such as molecular weight (MW), elemental composition and presence of specific functional groups (FG) can also constrain the search space. With enough constraints, one may arrive at a unique solution (chemical structure) that would satisfy all listed constraints, and in doing so, identify the unknown chemical. The drawback is that with the number of constraints, the performance of generative DL network drops dramatically. This decrease in performance was observed in our experiment. Designing valid chemicals that satisfy a large number of chemical property objectives remains a big challenge.

\begin{figure}[t]
\centering
\includegraphics[width=1\linewidth]{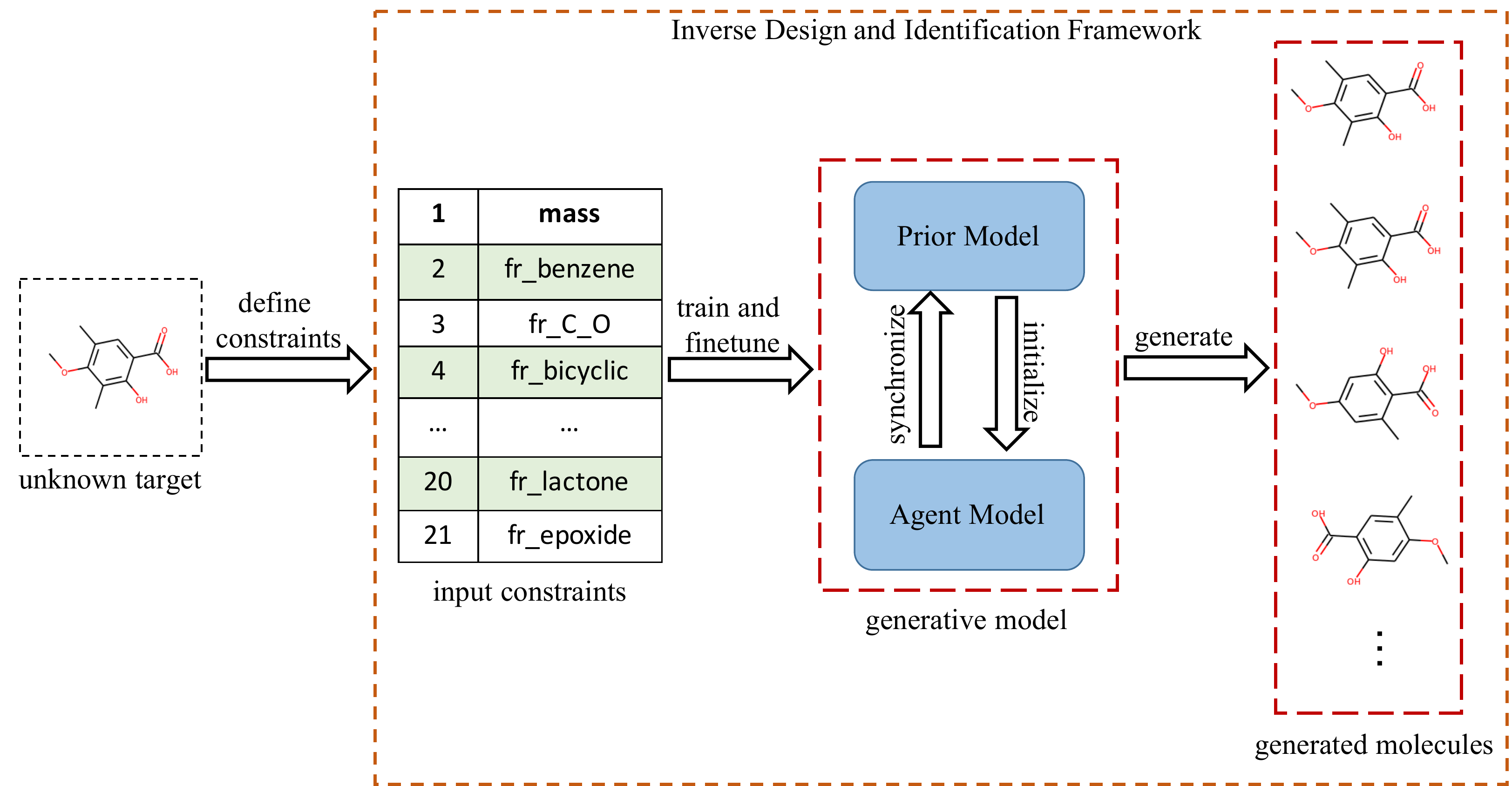}
\caption{Inverse chemical design and identification (identified constraint limit the chemical search space)}
\label{de-novo}
\end{figure}

Our approach for  inverse molecule design and identification is demonstrated in the Fig \ref{de-novo}. The target molecule is the unknown which needs to satisfy a list of input constraints that are the desired molecular properties. Note that ``constraints" and ``objectives" are exchangeable in this paper. We have two specific goals in this work: 1) propose a valid DL framework for inverse molecule design and identification; 2) increase the number of input constraints that can be satisfied by the model simultaneously, considering the high computational and time complexity in current strategies. The ultimate goal is to generate a molecule that is identical or close to the unknown target solely based on a sequence of constraints at high accuracy. We believe this work could have a significant contribution to the various fields mentioned above.
% Therefore, this is a ``blind'' study of identifying or generating an unknown chemical process. 

We formulate this molecule design and identification as a text generation and multi-constraint optimization problem. It is solved with the combination of RL and CL. Our contributions are as follows:
\begin{itemize}
    \item We developed training heuristics for multiple-objective (20+) RL using a modified curriculum training approach.
    \item We developed the first multiple-objective RL-based generative DL model for chemical identification of simple organic molecules that are relevant to biofuels applications.
\end{itemize}

\section{Related Work}

One of the first molecular generative models developed was based on conditional variational autoencoders (CVAE) \cite{gomez2018automatic,lim2018molecular}, which is used to convert molecules, represented as SMILES (a representation of molecules with ASCII alphabets), into a continuous vector representation. A major issue in early CVAE models is the low accuracy on generating valid SMILES, although recent work has made progress towards this goal\cite{dai2018syntax}. Other generative models have been reported based on generative RNN models, reinforcement learning (RL) \cite{olivecrona2017molecular}, and GANs \cite{kadurin2017cornucopia}, and these models tend to result in a higher proportion of valid SMILES than CVAE-based models.

All prior work thus far have been using generative models for chemical design, typically optimizing for a single objective \cite{popova2018deep,olivecrona2017molecular}. More recent work has included a few objectives, with typically no more than 5 properties optimized simultaneously\cite{lim2018molecular}. In addition, the use of generative models for chemical identification has yet to be reported. 

In this work, we combine RL with CL to extend the boundary of the number of property constraints that can be optimized in the molecule generation context. RL as a machine learning method that learns to conduct complex tasks with real environment interactions has received much attention in many domains, such as game playing\cite{mnih2015human} and robotic control\cite{lillicrap2015continuous}. RL also has been widely used in chemical design domains, such as molecule optimization with Deep Q-network\cite{zhou2019optimization} where actions are handcrafted molecular properties associated with the input constraints. In our work, we minimized human participation and considered the desired constraints as part of the RL reward function instead of merging it into actions. Our work is also related to the work of Popova et al\cite{popova2018deep} where a predictive model is trained separated from the generation model to forecast and bias the generation of new chemical structures towards the desired properties. Our model also consists of two parts: prior and agent. The prior is a general generative model, while the agent is used to optimize the constraints. In our effort, we particularly emphasized increasing the number of constraints that the model can optimize. Lastly, our work is also inspired by the related work  \cite{zhou2017optimizing,arus2019exploring} where Recurrent Neural Network is used to address the long-term dependencies between local molecular function groups.  To our knowledge, this is the first work that applied CL to the molecular inverse design problem. Previously, the application of CL to improve RL learning efficiency to deal with difficult or multiple tasks has achieved much success in other domains, such as game playing\cite{narvekar2017curriculum,narvekar2019learning} and robotic control \cite{florensa2017reverse}.

\section{Model Design}

In this section, we give an overview of the system design, including a \textbf{prior} model for generating valid SMILES, and an \textbf{agent} model fine-tuning through curriculum-based RL for optimizing multiple objectives/constraints.

\begin{figure} 
\centering
\includegraphics[width=0.5\linewidth]{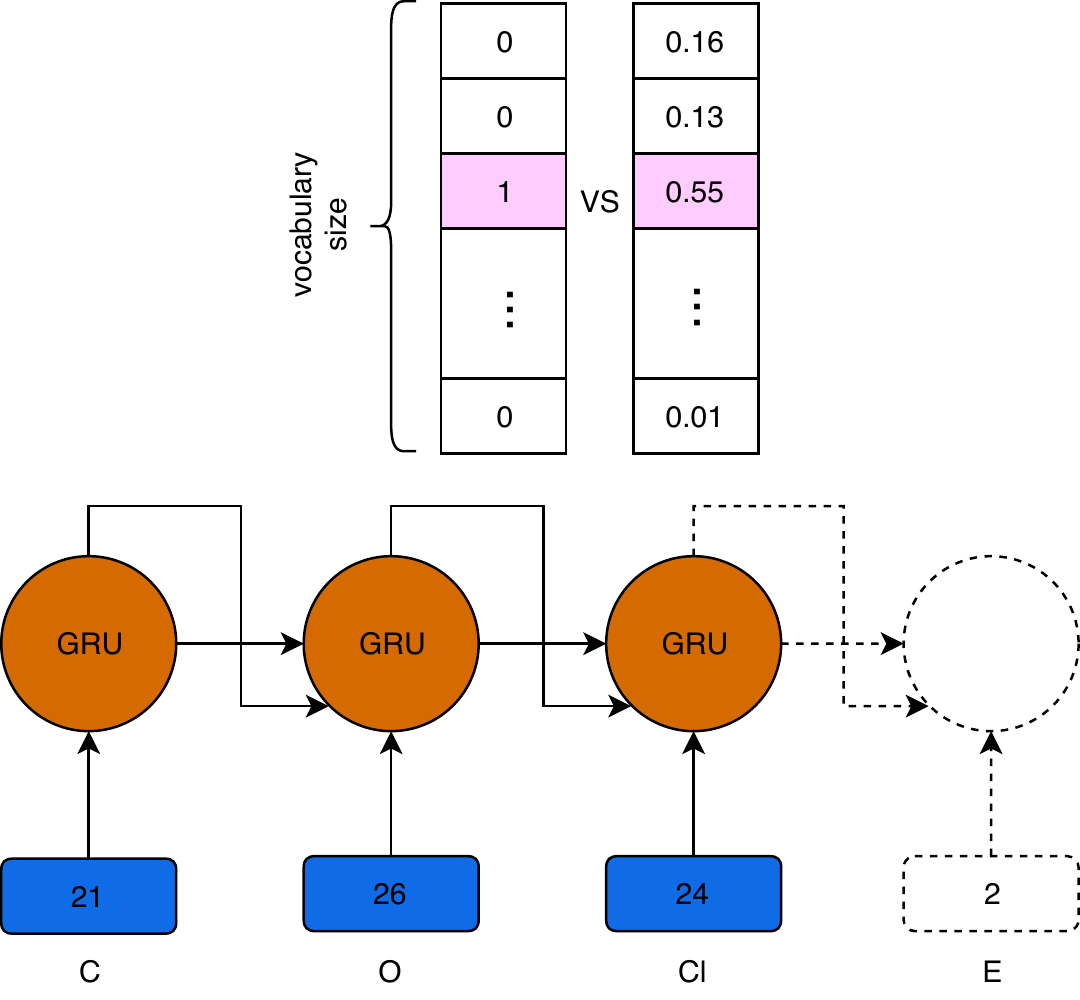}
\caption{RNN learning the SMILES syntax}
\label{GRU}
\end{figure}

\subsection{Prior Model}

The prior model is designed and trained to generate valid molecule SMILES sequences starting with only a start token feed. SMILES\cite{weininger1989smiles} is a molecule structure representation using ASCII format. Alphabets are used to represent atoms and molecular structures. For example ``cyclopropene" is written as ``C1=CC1''. We consider that the valid formulations of molecules are encoded in the SMILES sequences with the right chemical syntax. Previous research has demonstrated that the SMILES syntax can be learned efficiently with generative RNN models \cite{olivecrona2017molecular,popova2018deep}. Using a similar approach, we utilize an RNN(GRU)-based model to represent the prior model and learn the SMILES syntax. With the RNN layers, the training data is processed with an integrated loop and information flows from one step to the next while selectively remembering the past information, in this way, the long-term dependencies inside a sequence can be learned. 

The input for the prior model is the tokenized SMILES, and the output is the corresponding sequence-shifted one-hot vector. This means that when the first $t-1$ characters in a SMILES sequence are fed into the prior, the $t^{\text{th}}$ character is predicted but as a conditional probability distribution across the whole vocabulary, as shown in Fig \ref{GRU}. Assuming a single input/output pair is $\left (x, y \right)$ where $x$ represent the first $t-1$ characters and $y$ is the the $t^{\text{th}}$ character as well as the prediction target, and the $\hat{y}$ is the output of the prior model which is a conditional probability distribution $P(\hat{y}=c|x)$ where $c$ is the predicted character. The loss for a single generated SMILES is represented with the cross-entropy equation, shown as Eq \ref{cross_entropy}:

\begin{equation}
L(y, \hat{y};\theta)  = -\sum^{T}_{i=1}y(log(\hat{y}))
\label{cross_entropy}
\end{equation}

where $\theta$ are the trainable parameters in the prior model. $T$ is the fixed length of the prior model's output. Here we used 140 as the desired length with zero post-padding for SMILES outputs whose lengths are shorter. 

% The prior model is optimized by updating with the gradient descent on the L2-norm loss over a mini-batch of SMILES (the batch size is 128), as Eq \ref{sgd}

% \begin{equation}
% \begin{aligned}
% &\theta = \arg\min\sum_{(X,Y)}L^{2}(Y, \hat{Y}; \theta) \\
% &\text{where} \qquad \hat{Y} = \text{prior}(X; \theta)
% \end{aligned}
% \label{sgd}
% \end{equation}
% where $(X, Y)$ is one training batch with input and target and prior($\cdot$) is the prior model. 

% \subsection{Fine-Tuning Agent Model through Reinforcement Learning}
\subsection{Agent Model}
\begin{figure}
\centering
\includegraphics[width=.3\textwidth]{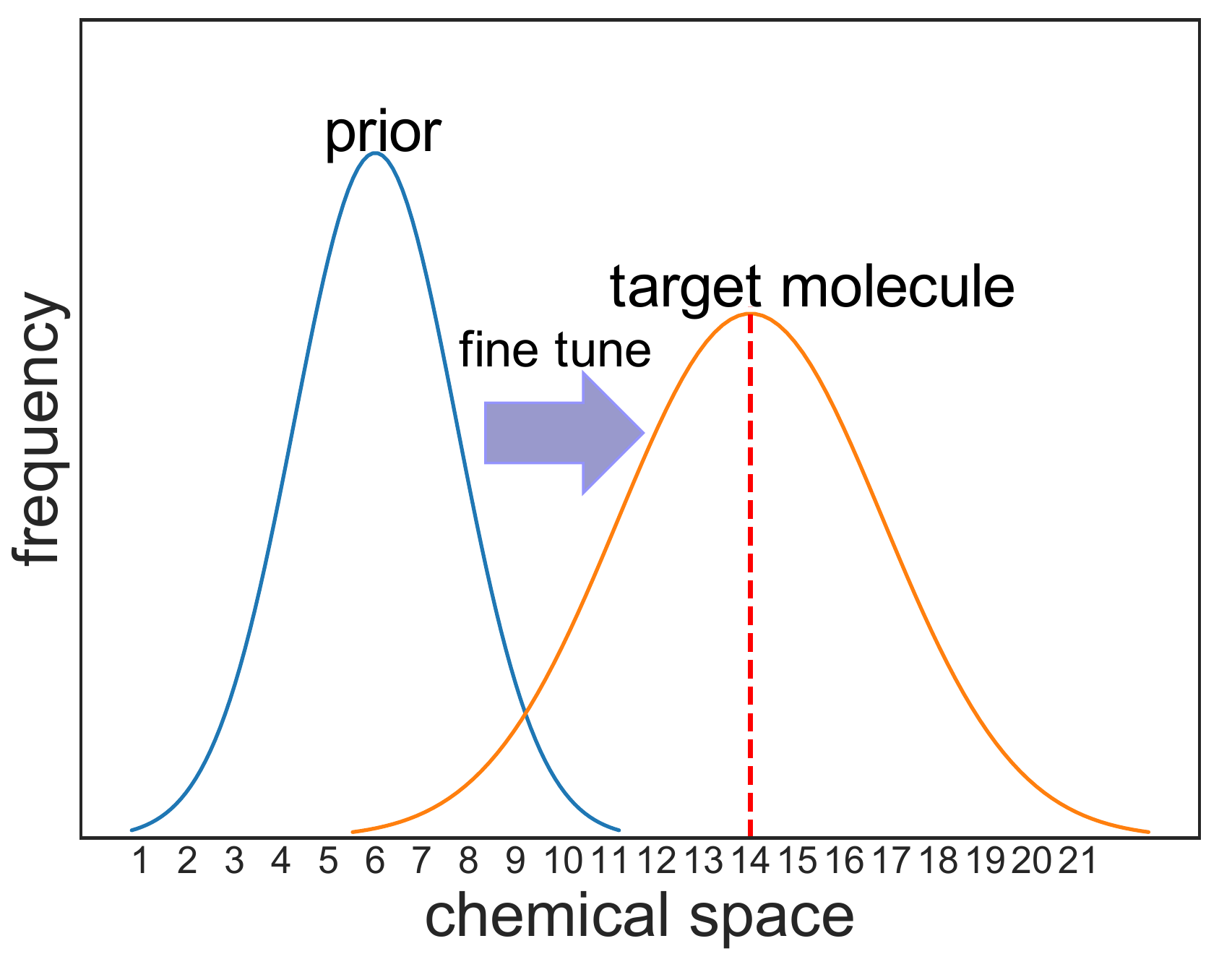}
\caption{Shift in generated SMILES distribution to fit the desired molecular distribution due to reinforcement learning}
\label{fine-tune distribution}
\end{figure}

After being trained with 150 million valid SMILES sequences (training batch size is 128) for 20 epochs, the prior model reaches $\sim 98 \%$ of SMILES generation validity. However, the generated SMILES do not necessarily satisfy any specific design or identification criteria, only the valid SMILES representations of plausible molecules. The desired SMILES that satisfy all constraints is nearly intractable for the prior model, as our model has a max output length of 140 and the vocabulary size considered (number of valid SMILES) is 87, which translates to a search space of up to $87^{140}$. Therefore, it is necessary to build an \textbf{agent} model to generate a higher fraction of desired SMILES more efficiently. The agent model is initialized from the prior model and tuned with multiple target molecular constraints. The agent model tuning is a distribution shifting process, as shown in Fig \ref{fine-tune distribution}. The SMILES sequences generated by the well-trained prior model follow a ``default" distribution over the chemical space, however, with the target molecular constraints, the distribution is expected to change. Therefore, tuning the agent model aims to make it able to generate SMILES with high validity as well as following a new target distribution. In this study, we use RL as the tuning approach.

\subsection{Tuning agent model with RL}
RL is a commonly used machine learning method for sequential decision-making problems, and it learns by interacting with the environment and maximizing the cumulative rewards. In this study, we show that the SMILES generative model optimization can be as studied as an RL problem, and it is modeled as a Markov Decision Process (MDP). The MDP can be represented as $\{\mathcal{S},\mathcal{A}, \mathcal{P}, \mathcal{R}, \rho \}$:
\begin{itemize}

\item The state space ($\mathcal{S}$) is a collection of any possible previous SMILES subsequence. For example, one state $s_{i}$ represent a sub-sequence $\{c_{0}, c_{1}, \cdots, c_{i}\}$ and $c$ is one alphabetic character. In our setting, states can have various dimensions instead of having a fixed one in other MDPs.

\item The action space ($\mathcal{A}$) is the entire vocabulary used to represent SMILES, and each action is one alphabet(character) in the vocabulary. Therefore, the vocabulary's size is the action space's size.

\item The transition model ($\mathcal{P}$) represents the stochasticity of the generative model. Given a current state $s_{i-1}=\{c_{0}, \cdots, c_{i-1}\}$ and an action $a_{i-1}$, the probability of next state $s_{i}=\{c_{0}, \cdots, c_{i}\}$ is $\mathcal{P}(s_{i}|s_{i-1}, a_{i-1})$ and this is the output of the generative model. It also can be used to balance the RL exploration and exploitation: sampling an action following this probability distribution is the exploration and taking the one action with the highest probability is the exploitation. 

\item The reward function ($\mathcal{R}(y) \rightarrow r \in \mathbb{R}$) is aggregated and measures both the sequence validity and the distance to the desired SMILES w.r.t the desired constraints:

\begin{equation}
\mathcal{R}(y) = L(y_{a}, \hat{y}_{p}) + \mathcal{C}(y_{a})
\label{aggregate_r}
\end{equation}

The first term $L(y_{a}, \hat{y}_{p})$ represents the syntax validity of a sequence $y_{a}$ generated by the agent model compared to the pretrained prior model. $\hat{y}_{p}$ is the output of the prior model given the same input with the agent model, and it's the cross-entropy loss calculated with Eq \ref{cross_entropy}. $\mathcal{C}(\cdot)$ is the \textbf{constraint score} measuring how much  $y_{a}$ satisfies the desired constraints (we'll introduce details in the following sections). The reward function is defined at the SMILES sequence level so that a reward is only provided at the end of each RL trajectory. Each trajectory is a SMILES sequence that is iteratively predicted by the agent model until a terminal token is reached. 

\item The initial state $\rho$ is the same initial token ``G" for all SMILES sequences. 

\end{itemize}

The agent model follows the RL policy $\pi(a_{i}|s_{i})$ to make a valid prediction and the goal is to maximize the objective function, $\eta(y_{a}; \theta_{a})$, shown as Eq \ref{tuning_loss}.

\begin{equation}
\begin{aligned}
\eta(y_{a}; \theta_{a}) &= \mathcal{R}(y_{a}) - L(y_{a}, \hat{y}_{a}; \theta_{a}) \\ & = \mathcal{C}(y_{a}) + (L(y_{a}, \hat{y}_{p}; \theta_{a}) - L(y_{a}, \hat{y}_{a};\theta_{a})) 
\end{aligned}
\label{tuning_loss}
\end{equation}

The agent model's trainable paramters ($\theta_{a}$) are optimized (as Eq \ref{param_update} ) by updating using gradient ascent with the objective function, as Eq\ref{param_update}

\begin{equation}
    \theta_{a} \leftarrow \theta_{a} + \bigtriangledown_{\theta}\eta(y_{a}; \theta_{a})
\label{param_update}
\end{equation}

The difference of cross-entropy loss is an inverse Kullback–Leibler (KL)-divergence \cite{kullback1951information}, shown in Eq \ref{KL}. Maximizing the inverse KL divergence is to minimize the difference of agent model's policy and the pior model's policy and thus maintain the valid SMILES syntax. Here the well-trained prior is used as the target since it has near 100\% generative validity. 

\begin{equation}
\begin{aligned}
-D_{KL}(y_{a}|y_{p}) & = L(y_{a}, \hat{y}_{p}) - L(y_{a}, \hat{y}_{a})\\
& = \sum_{i}y_{a, i}\log \frac{\hat{y}_{a, i}}{\hat{y}_{p, i}} \propto \sum_{i} \hat{y}_{a, i} \log \frac{\hat{y}_{a, i}}{\hat{y}_{p, i}} 
\end{aligned}
\label{KL}
\end{equation}

In related work, when there is more than 1 desired constraint, a naive (baseline) approach is to use an equal weighted sum as the constraint score\cite{lim2018molecular}, as Eq \ref{baseline_reward}.
\begin{equation}
\mathcal{C}(y) = \frac{1}{k}\sum_{i=0}^{k}\mathcal{C}_{i}(y)
\label{baseline_reward}
\end{equation} 
where $k$ is the number of constraints. However, the performance of this naive approach is unlikely to scale well to multiple constraints. We hypothesize that part of the issue is caused by the equal weights, where the influence from different directions are unequal or even cancelling out one another. Another issue is the inability to find a good local minimum solution that satisfies all constraints in an extensive search space across multiple objectives. To address these limitations, we develop heuristics inspired from CL.

\subsection{Curriculum Learning (CL)}

\begin{figure}[t]
\centering
\includegraphics[width=.4\textwidth]{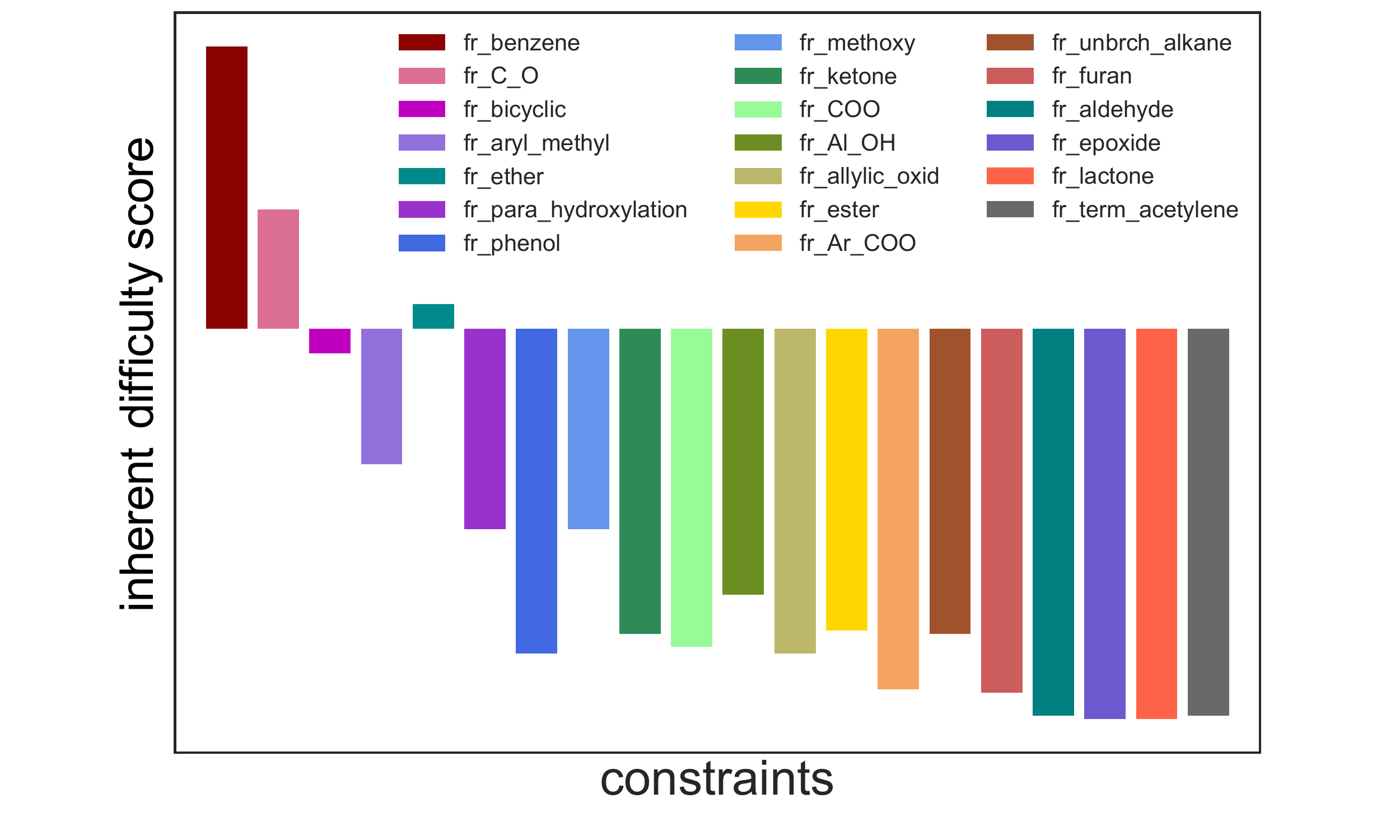}
\caption{Average inherent reward score of the 20 FG constraints}
\label{20constraints}
\end{figure}

CL is a strategy for multi-task learning, which draws parallels to human knowledge. A sequence of subtasks is ranked in difficulty ascending order and conducted within different phases\cite{bengio2009curriculum}. Its success in learning performance improvement can be seen in many domains, such as learning language modeling and pattern recognition \cite{bengio2009curriculum,graves2017automated}. The difficulty level of subtasks is usually defined with respect to the training data. In this study, we design a measurement (difficulty score) to differentiate task difficulties and tune the agent model to satisfy a larger number of constraints with multiple tuning phases. The goal for the agent model is to generate valid SMILES satisfying a molecular mass constraint and 20 fingerprinting (FP) objectives. A fingerprinting objective is a quantitative molecular structural or functional property.

\begin{figure}
\centering
\includegraphics[width=.45\textwidth]{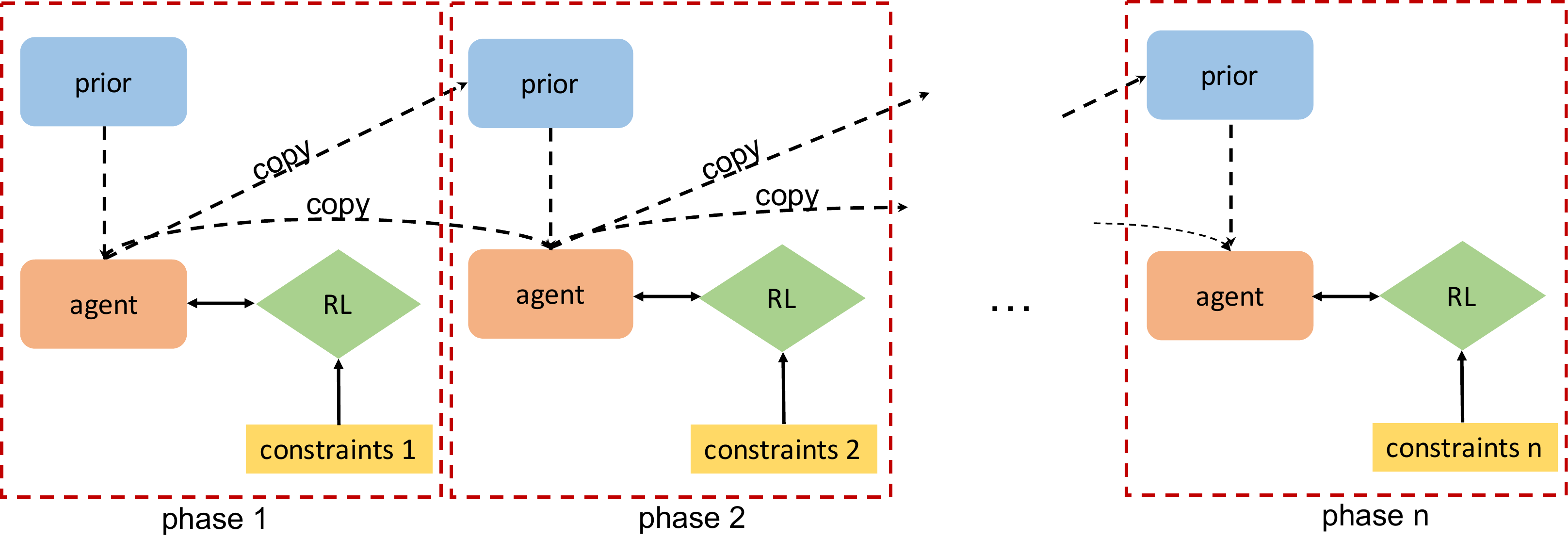}
\caption{Agent model is trained with RL and curriculum learning}
\label{curriculum_train}
\end{figure}

The difficulty score is to measure the difficulty of each constraint. For one constraint $j$, the difficulty score $\mathcal{D}_{j}$ is the percentage of it being captured in the prior distribution with the pretrained prior model, as calculated with Eq\ref{baseline}:

\begin{equation}
\mathcal{D}_{j} = \frac{1}{\left | Y \right |}\sum_{y \in Y}F_{j}(y) \qquad F_{j}(y) \in \{-1, 1\} 
\label{baseline}
\end{equation}

% The difficulty score ($F_{k}$) for target constraint $k$ is calculated using Eq \ref{baseline} 
where $F_{j}(y)$ is 1 if the sequence $y$ contains the target constraint, otherwise, -1. $\left | Y \right |$ is the size of SMILES test dataset. Using 1000 SMILES string generated by the prior model, we report the average inherent difficulty score $\mathcal{D}$ as shown as Fig \ref{20constraints}. The inherent difficulty score can be interpreted as the ``default'' probability that a certain constraint is satisfied. For example, the FG constraint for benzene is about 0.8, which means that satisfying this constraint is relatively easy since it can be generated 80\% of the time. With the difficulty score ($\mathcal{D}$), the constraints can be ranked from easiest to the most difficult. Progressively more difficult tasks are designated by having a greater divergence from the inherent distribution relative to the prior model. 

In this work, we propose a novel approach of combining CL heuristics in a RL context. We hypothesize that the agent model can be more effectively trained to satisfy multiple constraints using this approach. This RL-based CL process is demonstrated as Fig \ref{curriculum_train}. There are multiple curriculum training phases. In each phase, the agent model is initialized from the prior model in the beginning, and the prior model is synchronized with the agent model at the end of each phase for memorizing the previously learned constraints. As one progresses through phases, new constraints will be merged with the previous ones as part of the RL reward for agent model tuning. The constraint score, $\mathcal{C}(y)$ at phase $k$ is updated accordingly as:

\begin{equation}
\mathcal{C}_{k}(y) = (1-w)\mathbb{E}_{j=1, \cdots, k-1}(\mathcal{C}_{j}(y)) + w \mathcal{C}_{k}(y)
\label{phase_reward}
\end{equation}
where $w$ is a constant weight to balance the previous fine-tuning constraints and the newly-added one(s).  

\section{Method and Experiment}

\subsection{Dataset and Pre-processing}
\label{data_details}

We used the ChEMBL \cite{gaulton2016chembl} database as our training dataset that has 1.5 million syntactically valid SMILES. In the ChEMBL vocabulary, the complex characters `Cl' and `Br' are singularized to tokens `R' and `L'.  A start token (`G') and a termination token (`E') are also added. With the start and terminal token, the vocabulary size of unique alphabets and symbols used in SMILES training SMILES dataset is 87. The input to the prior is the tokenized SMILES, for example, a molecule '[nH]1cnc2cncnc21' is wrapped with token 'G' and 'E' firstly, then alphabetic characters are mapped to a integer vector as [1,78,10,82,83,82,11,82,83,82,83,82,11,10,2]. The output is the one-hot vectors of the same SMILES with offset by +1. Post zero padding is applied on the encoded input to be obtained a unified length of 140.

\subsection{Training Prior Model}

The prior model network architecture starts with a single embedding layer, followed by a stacked 3-layer GRU with 512 cells per layer, and ending with a single fully-connected layer. The model was trained on mini-batch (size 128) for 20 epochs with early stopping. RMSprop is used as the optimizer with initial learning rate 0.001 and gradient values are clipped to $ \left [-3, 3\right ]$. The prior model generates the next token with a given input and advances time by one step and recurrently extends the input with the newly generated token until it generates the terminal token `E'. After training, the prior model is able to generate around 98\% valid SMILES.

\subsection{Prospective Test Set}

For the purpose of prospectively identifying unknown chemicals, we limit the search space to biomass-derived liquid relevant chemical space, which is defined as low molecular weight (MW) (< 200g/mole) organic compounds, containing only elements C, H and O. Using these criteria, we extracted a subset of 25,901 relevant chemicals from the original the ChEMBL database. We then used RDKit\cite{landrum2006rdkit} to compute descriptors pertaining to molecular weight (MW) and functional groups represented as FG constraints to simulate experimental characterization data from Mass Spectrometry(MS) and Nuclear Magnetic Resonance(NMR) sources. These descriptors are used as the constraints in our reward function.

\subsection{Designing the Constraint Score}
The constraint score $\mathcal{C}_{\text{mw}}$ for molecule mass (MW) is adapted from \cite{neil2018exploring}:

\begin{equation}
\mathcal{C}_{\text{mw}}(y) = \max(-1, \frac{1}{10^4}(x - \text{MW}(y))^{2}+1)
\end{equation}
where $\text{MW}(\cdot)$ calculates the real mass of a given SMILES. The other 20 FG constraints can be represented as $\left \{ f_{1}, f_{2}, \cdots, f_{20} \right \}$ and $f_{i} \in \left \{ \text{False}, \text{True} \right\}$, where ``True" means this FG is desired to be present in a given SMILES and vice versa.

\begin{equation}
\mathcal{C}_{f_{i}}(y) = \left\{\begin{array}{r@{}l@{\qquad}l}
    1 & \qquad \text{if} \quad f_{i} \in y \quad \text{and} \quad f_{i} = \text{True}\\
    1 & \qquad \text{if} \quad f_{i} \notin y \quad \text{and} \quad f_{i} = \text{False} \\
    -1 & \qquad \text{if} \quad f_{i} \in y  \quad \text{and} \quad f_{i} = \text{False} \\
    -1 & \qquad \text{if} \quad f_{i} \notin y \quad \text{and} \quad f_{i} = \text{True}
  \end{array}\right.
  \label{reward_function}
\end{equation}

The MW and presence of FG groups $f$ can be easily evaluated with a python package RDKit\cite{landrum2006rdkit}. MW approximately controls the length of the SMILES string, which is a significant factor in determining the extensiveness of chemical space search. Therefore, we designated a constant weighting factor ($\beta =0.5$) in the curriculum reinforcement reward function for MW. Combined with Eq\ref{phase_reward}, the constraint score ($\Bar{\mathcal{C}}$) at phase $k$ is finalized as:

\begin{equation}
\Bar{\mathcal{C}}_{k}(y) = \beta\mathcal{C}_{\text{mw}}(y) + (1-\beta) \mathcal{C}_{k}(y)
\label{final_constraint}
\end{equation}

\subsection{Tuning Agent Model}
We use the machine learning approach done in previous work as the baseline where the fine-tuning loss objective is the mean value over all constraints with equal weights and the model is fine-tuning with all constraints together without subtasks, shown as Eq \ref{baseline_reward}

% \begin{equation}
% \mathcal{C}(y) = \mathbb{E}(\mathcal{C}_{mw} + \frac{1}{n}\sum^{n}_{i=1}\mathcal{C}_{i}(y))
% \label{baseline_reward}
% \end{equation}

\begin{equation}
\Bar{\mathcal{C}}_{\text{baseline}}(y) = \mathbb{E}(\mathcal{C}_{mw} + \mathbb{E}_{f_{i}}\mathcal{C}_{f_{i}}(y))
\label{baseline_reward}
\end{equation}

\begin{table*}
\centering
\caption{Agent Model Fine-Tuning Approaches}
\begin{tabular}{|l|l|l|l|l}
\cline{1-4}
\multicolumn{1}{|c|}{Method} & \multicolumn{1}{c|}{No. of bins} & \begin{tabular}[c]{@{}l@{}}No. of retrains \\ (at each phase)\end{tabular} & No. of phases & \\ \cline{1-4}
Baseline & 1 & 0 & 1 \\ \cline{1-4}
Retrain Fine-tuning (RF\_2) & 1 & 1 & 2 \\ \cline{1-4}
Curriculum Fine-tuning (CF\_2) & 2 & 0 & 2     \\ \cline{1-4}
Curriculum Retrain Fine-tuning (CRF\_2) & 2 & 1 & 2\\ \cline{1-4}
Retrain Fine-tuning (RF\_4) & 1 & 1& 4          \\ \cline{1-4}
Curriculum Fine-tuning (CF\_4)& 4 & 0& 4       \\ \cline{1-4}
Curriculum Retrain Fine-tuning (CRF\_4) & 4& 1 & 4\\ \cline{1-4}
Retrain Fine-tuning (RF\_6)& 1  & 1& 6         \\ \cline{1-4}
Curriculum Fine-tuning (CF\_6)& 6& 0 & 6       \\ \cline{1-4}
Curriculum Retrain Fine-tuning (CRF\_6) & 6& 1 & 6\\ \cline{1-4}
\end{tabular}
\label{method_table}
\end{table*} 

With CL methods, we investigated 3 different approaches to fine-tune the agent model with RL: 
\begin{enumerate}[leftmargin=3\parindent]
    \item Retrain-based Fine-tune (RF) 
    \item Curriculum Fine-tune (CF)
    \item Curriculum Retrain-based Fine-tune (CRF)
\end{enumerate}

RF is an extension from the naive baseline with CL for a better comparison. Instead of a one-time tuning, it retrains the model with multiple phases. However, the reward function (as in Eq \ref{baseline_reward}) remains the same in all phases. In the CF method, the model is trained gradually with new constraints sequentially added at each phase without model retraining. Compared to the RF, CF is designed to prove the significance of separating the constraints into different difficulty levels. One observation is that the agent model's validity drops dramatically with too many tuning phases because each tuning with constraints is conducted while sacrificing of the generation validity.  To balance the model performance and CL, we group the constraints into multiple bins according to their difficulty so that the number of tuning phases is reduced. A new bin of constraints is added at each phase. For example, if we set 2 bins, there are about 10 constraints in each bin, and the first bin has easier difficulty than the second bin. The CRF method combines and alternates between both RF and CF approaches. The constraints are introduced into the reward function in a curriculum-based manner, but after each phase, the agent model is retrained with the same reward function one more time. A summary of the various methods settings with a different number of constraint bins (represented as the ``\_n" where n is the number of bins) is summarized in Table \ref{method_table}.

As with the baseline approach, the agent is updated with mini-batch gradient descent with size 128, and gradients are clipped to $\left [ -3, 3 \right ]$. Each phase also has an early stopping algorithm, where training terminates if the reward does not improve after 50 iterations, and the last checkpoint is saved.

\section{Experiment Results and Analysis}

\begin{figure*}[h!]
\centering
\subfigure[top 5 constraint scores]{
\includegraphics[width=0.4\linewidth]{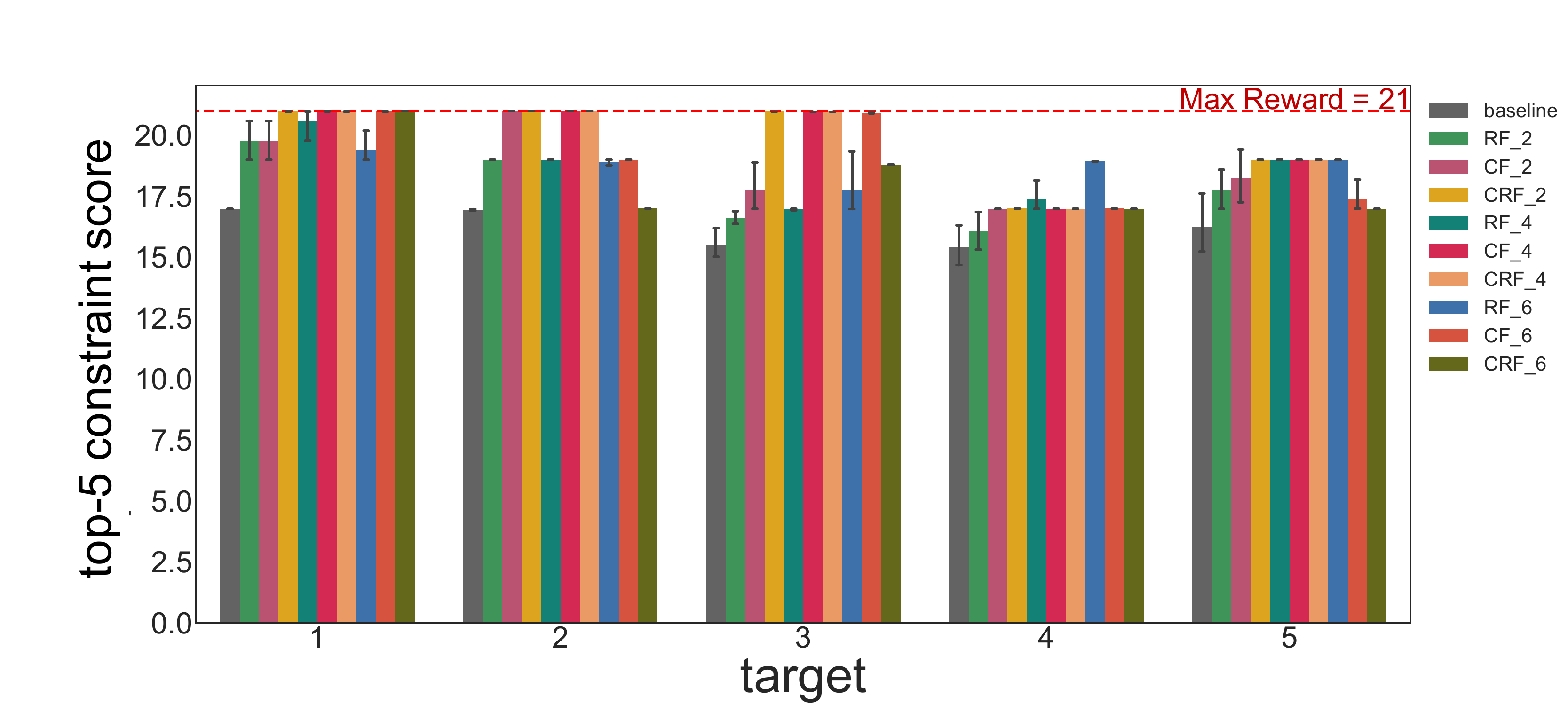}
\label{01_score}}
\subfigure[top 5 average similarities]{
\includegraphics[width=0.4\linewidth]{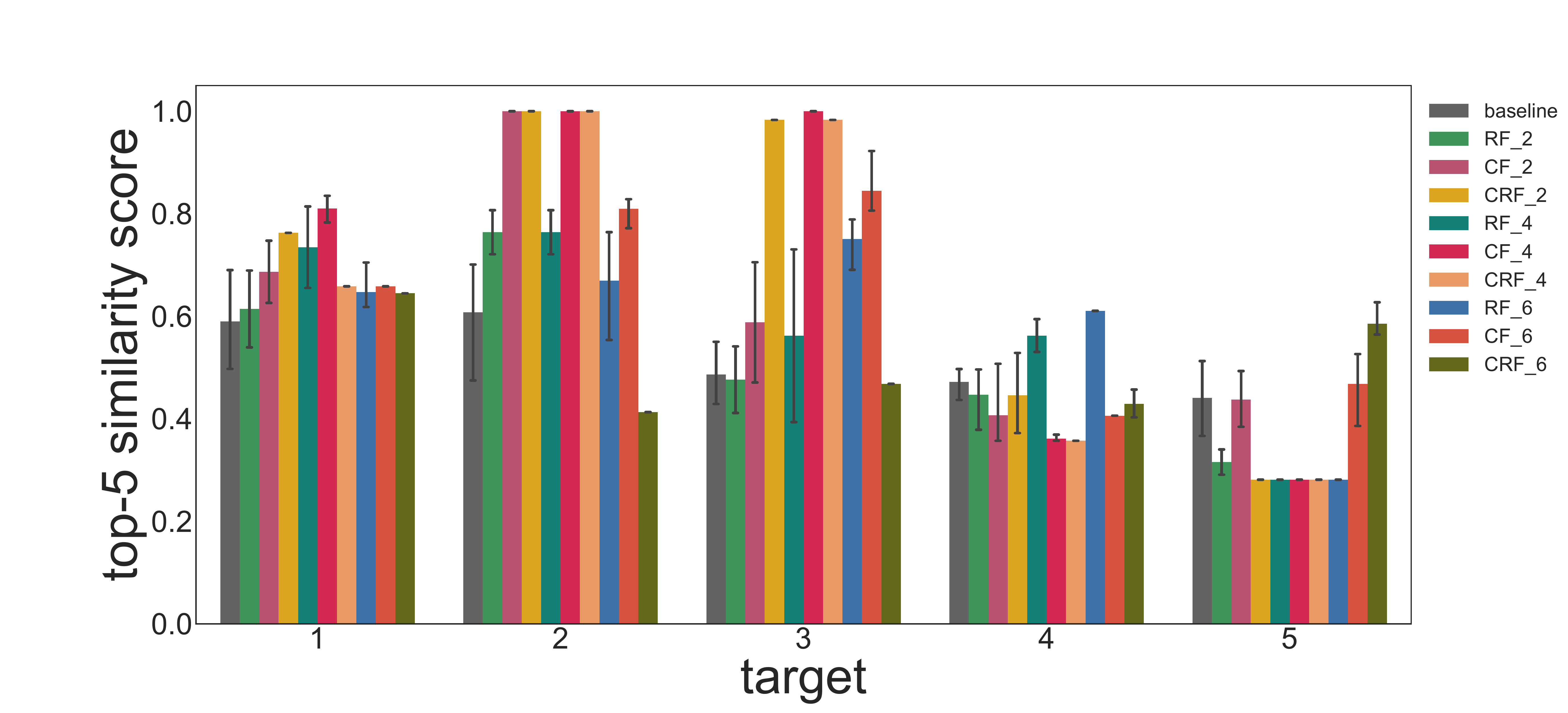}
\label{01_sim}}

\subfigure[refined top 5 constraint scores]{
\includegraphics[width=0.4\linewidth]{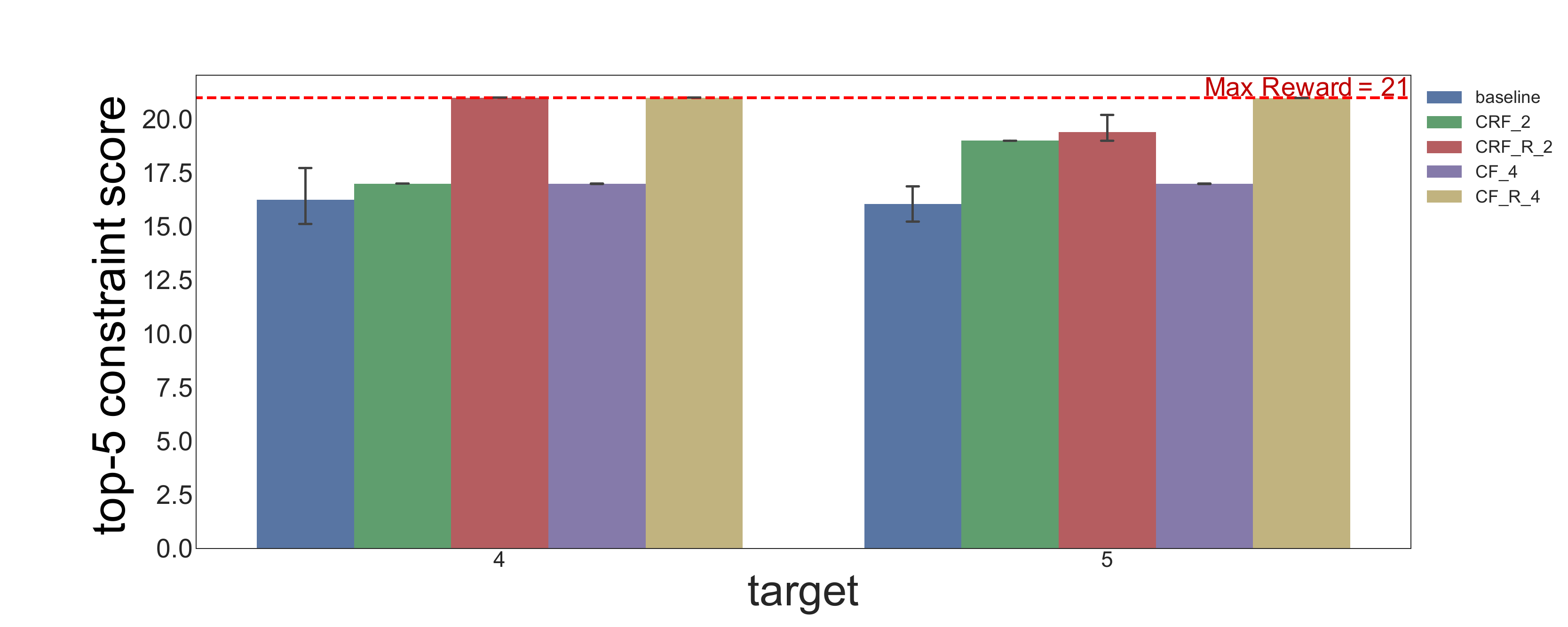}
\label{imp_score}}
\subfigure[refined top 5 average similarities improvement]{
\includegraphics[width=0.4\linewidth]{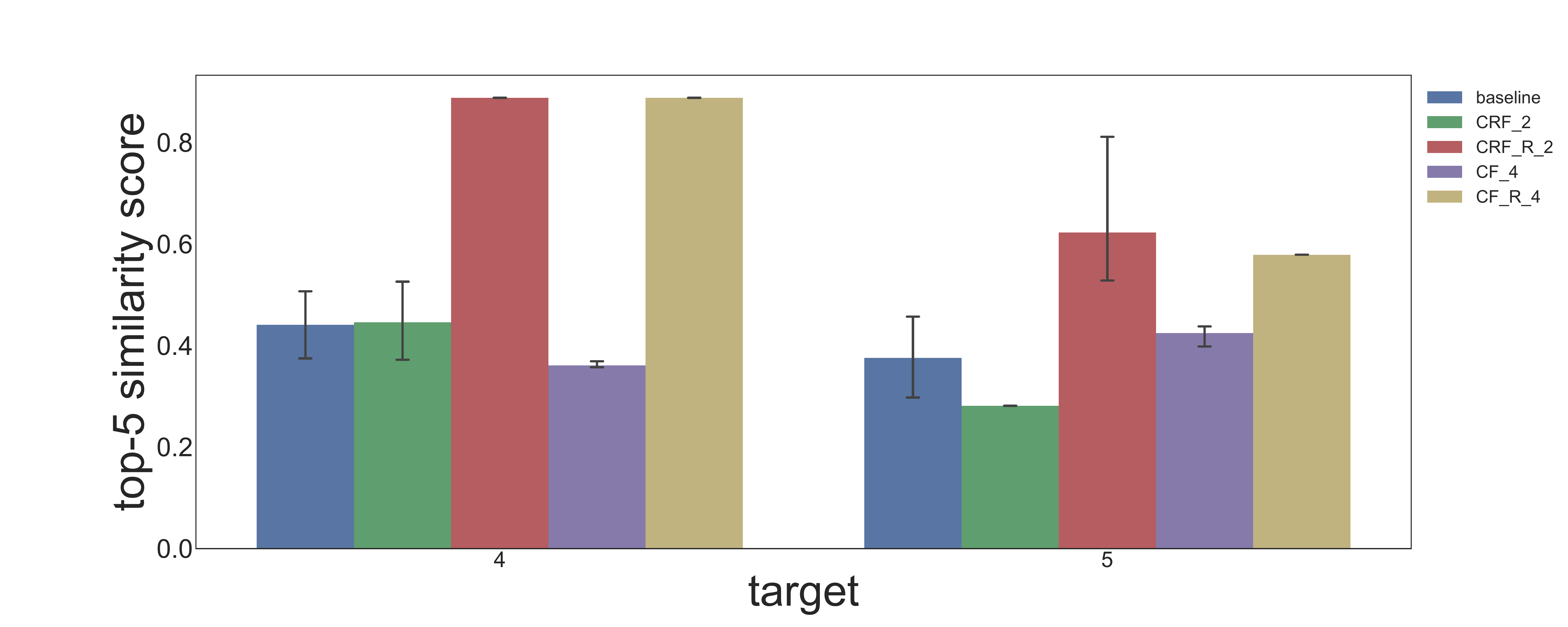}
\label{imp_sim}}

\caption{Agent fine-tuning performance across 5 chemical targets (as Fig \ref{01_score} and \ref{01_sim}). Adaptive weight refinement of reward function improves results (as Fig \ref{imp_score} and Fig \ref{imp_sim})}
\label{result}
\end{figure*}

\begin{figure}[t]
\centering
\includegraphics[width=0.9\linewidth]{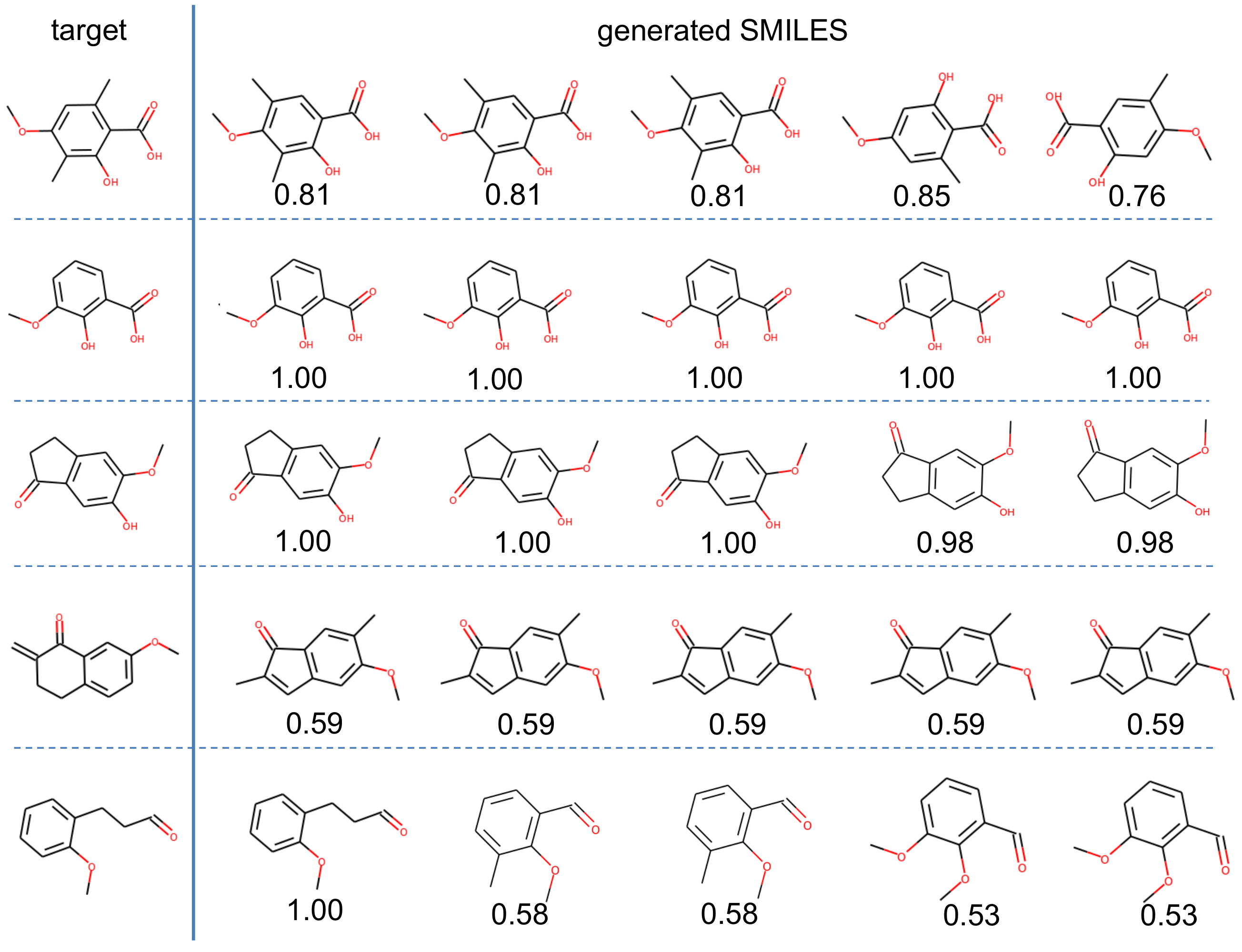}
\caption{Agent generated molecules comparing with targets}
\label{visua_smile}
\end{figure}

We randomly picked 5 biofuel-relevant molecules as the targets (shown in Fig \ref{visua_smile}) and formed the corresponding constraints as the framework input. We evaluated the performance of CF with 2, 4 and 6 constraints bins. Controlling for total training time and number of updates, we also evaluated the RF and CRF with 2, 4, and 6 retraining phases. In total, the baseline method and 9 variations CF/RF/CRF were tested for each of the 5 selected target molecules.

We evaluate the final performance of each model with two metrics: 1) the constraint score $\mathcal{C}$ over all desired constraints; 2) the similarity between the generated molecule and the target molecule with the Tanimoto distance \cite{tanimoto1958elementary} which also can be calculated with RDKit. The similarity score is used only as a post ad-hoc evaluation once the model has been well trained. It has no bearing on the model's constraint score function. 

The performance measured with the total constraint score is shown in Fig \ref{01_score}. The total constraint score is calculated with 256 randomly sampled SMILES generated by the tuned agent model. For each target and each method, all generated SMILES are sorted based on the constraint score in a descending order. We selected the 5 SMILES with the highest scores to show (named as top-5 constraint score). An underlying assumption is that a model that generates SMILES with high constraint scores (thus satisfying most if not all constraints) will exhibit high similarity to the target molecule. The corresponding similarity compared to the target molecules is also shown in Fig \ref{01_sim}.

The proposed approaches all outperform the baseline approach in the top-5 constraint score, with several models achieving close to the maximum reward score of 21. Overall, the approaches with curriculum fine-tuning have consistent good performance, and there is no significant difference between curriculum fine-tuning with model retrain (CRF) and without retrain (CF). With 4 constraint bins (5 constraints in each), the agent model showed better and robust performance compared to other bin numbers. This observation matches the previous related discovery that 5 is the upper bound of constraints that are equally weighted as the RL reward for a generative model under a good performance condition. On the other hand, the approaches with 5 bins outperform the ones with more bins, and it proved that the number of phases in CL matters regarding the model performance as well. When evaluated against similarity, our proposed methods all perform better than baseline on the first three targets. 
% We also visualize the generated SMILES and compare to the target molecule (shown in Fig \ref{visua_smile}), the exact molecules were identified for targets 2 and 3.

However, there is an apparent disconnection for targets 4 and 5, as even though the top-5 reward score is higher than the baseline, that is not translated to a higher similarity score. To explain this, we have two hypotheses: (1) the model gets stuck at a local optima trained by RL during the early phases, and hence it was unable to arrive at a global optima at the last phase, i.e., the agent learns early objectives ``too-well'' and is unable to learn subsequent objectives; (2) certain constraints are mutually exclusive. For example, we observed that the existence of constraints "fr\_allylic\_oxid" limits the appearance of  "fr\_aldehyde" in the same molecule. The latter hypothesis is an inherent disadvantage and may be addressed by human experts. In this work, we focus on addressing the first one.

To avoid the model getting stuck at a "bad'' local optima in early phases, we explored a heuristics to reshape the constraint function $\Bar{\mathcal{C}}'$ so that some difficult constraints can be addressed after the curriculum tuning, as shown in Eq \ref{refine_eq}. 
\begin{equation}
\Bar{\mathcal{C}}' = \mathbb{E}(\mathcal{C}_{mw}, \mathcal{C}_{k}, \mathbb{E}_{-f_{k}}(\mathcal{C}_{f_{1}}, \cdots,\mathcal{C}_{f_{k-1}},\mathcal{C}_{f_{k+1}}))
\label{refine_eq}
\end{equation}

where $\mathcal{C}_{f_{k}}$ is the weakly-learned constraint and it is picked by comparing with a pre-defined threshold ($\xi$) w.r.t the constraint score ratio compared to the maximum value and here $\xi = 0.5$:
\begin{equation}
k \leftarrow \arg_{i \in [1, 20]}(\frac{\mathcal{C}_{f_{i}}}{\mathcal{C}_{\max}}< \xi)
\end{equation}

$\mathcal{C}_{\max}$ is maximum constraint score, here is 21. If there are more than 1 weakly learned constraints, the average value over all weakly-learned constraints is used instead. Using this heuristic, we refined the CRF\_2 and CF\_4 models on targets 4 and 5 with the reshaped constraint function $\Bar{\mathcal{C}}'$, and add  `\_R' at the middle to notate this is a refinement approach. For example, 'CF\_R\_4' representing the refined CL heuristics with 4 bins. Using this refined approach, the resulting top-5 constraint score achieves near the maximum value (shown in Fig \ref{imp_score}). Furthermore, we also observed that similarity to the targets was significantly improved (Fig \ref{imp_sim}). 

Lastly, we visualized the top-5 generated SMILES, and compare to the target molecule (summarized in Fig \ref{visua_smile}) with the similarity labeled below. The generated SMILES correctly identified the unknown target 2, 3 and 5. Target 1 identification was sufficiently close, as structural isomers of the target was generated. We note that the prediction of target 4 is not the same as its target, and we hypothesize that this is because there may not be sufficient or appropriate constraints for that structure, and so the search space remains large (i.e. it is an undetermined search problem).

\section{Conclusion}
In conclusion, building on earlier work \cite{olivecrona2017molecular,segler2017generating,neil2018exploring} of RL-based generative RNN model, we enhance the functionality of such methods to include multiple-objectives in the reward function. In this study, we proposed and evaluated several curriculum-based reinforcement learning heuristics, and showed that at least 21 different objectives that incorporate both MW and FG constraints could be simultaneously optimized.

The naive approach of an equally weighted reinforcement learning heuristic with a single-pass training achieves poor results, both in its ability to achieve maximum reward score as well as similarity to the unknown target. In general, our results indicate that a curriculum-learning approach consistently outperforms baseline. However, there are complications associated with having too many learning phases, possibly because it gets trapped in a bad local optimum in the beginning, and so is unable to learn latter objectives effectively. To address these limitations, we developed further heuristics that adaptively overweight unlearned objectives. Our results show that for all five molecules tested, and maximum top-5 reward score can be obtained in all 5 cases tested. In addition, a maximum top-5 reward score also translates well to chemical similarity, as 4 out of 5 molecules were identified correctly. Lastly, the curriculum-learning based heuristics developed in this work is also a significant improvement to the baseline model, which does not even identify a single molecule correctly.

\bibliography{ref} 

\begin{thebibliography}{}

\bibitem[\protect\citeauthoryear{Ar{\'u}s-Pous \bgroup et al\mbox.\egroup
  }{2019}]{arus2019exploring}
Ar{\'u}s-Pous, J.; Blaschke, T.; Ulander, S.; Reymond, J.-L.; Chen, H.; and
  Engkvist, O.
\newblock 2019.
\newblock Exploring the gdb-13 chemical space using deep generative models.
\newblock {\em Journal of cheminformatics} 11(1):20.

\bibitem[\protect\citeauthoryear{Bengio \bgroup et al\mbox.\egroup
  }{2009}]{bengio2009curriculum}
Bengio, Y.; Louradour, J.; Collobert, R.; and Weston, J.
\newblock 2009.
\newblock Curriculum learning.
\newblock In {\em Proceedings of the 26th annual international conference on
  machine learning},  41--48.
\newblock ACM.

\bibitem[\protect\citeauthoryear{Bingol}{2018}]{bingol2018recent}
Bingol, K.
\newblock 2018.
\newblock Recent advances in targeted and untargeted metabolomics by nmr and
  ms/nmr methods.
\newblock {\em High-throughput} 7(2):9.

\bibitem[\protect\citeauthoryear{Boiteau \bgroup et al\mbox.\egroup
  }{2018}]{boiteau2018structure}
Boiteau, R.~M.; Hoyt, D.~W.; Nicora, C.~D.; Kinmonth-Schultz, H.~A.; Ward,
  J.~K.; and Bingol, K.
\newblock 2018.
\newblock Structure elucidation of unknown metabolites in metabolomics by
  combined nmr and ms/ms prediction.
\newblock {\em Metabolites} 8(1):8.

\bibitem[\protect\citeauthoryear{Dai \bgroup et al\mbox.\egroup
  }{2018}]{dai2018syntax}
Dai, H.; Tian, Y.; Dai, B.; Skiena, S.; and Song, L.
\newblock 2018.
\newblock Syntax-directed variational autoencoder for structured data.
\newblock {\em arXiv preprint arXiv:1802.08786}.

\bibitem[\protect\citeauthoryear{Florensa \bgroup et al\mbox.\egroup
  }{2017}]{florensa2017reverse}
Florensa, C.; Held, D.; Wulfmeier, M.; Zhang, M.; and Abbeel, P.
\newblock 2017.
\newblock Reverse curriculum generation for reinforcement learning.
\newblock {\em arXiv preprint arXiv:1707.05300}.

\bibitem[\protect\citeauthoryear{Gaulton \bgroup et al\mbox.\egroup
  }{2016}]{gaulton2016chembl}
Gaulton, A.; Hersey, A.; Nowotka, M.; Bento, A.~P.; Chambers, J.; Mendez, D.;
  Mutowo, P.; Atkinson, F.; Bellis, L.~J.; Cibri{\'a}n-Uhalte, E.; et~al.
\newblock 2016.
\newblock The chembl database in 2017.
\newblock {\em Nucleic acids research} 45(D1):D945--D954.

\bibitem[\protect\citeauthoryear{G{\'o}mez-Bombarelli \bgroup et
  al\mbox.\egroup }{2018}]{gomez2018automatic}
G{\'o}mez-Bombarelli, R.; Wei, J.~N.; Duvenaud, D.; Hern{\'a}ndez-Lobato,
  J.~M.; S{\'a}nchez-Lengeling, B.; Sheberla, D.; Aguilera-Iparraguirre, J.;
  Hirzel, T.~D.; Adams, R.~P.; and Aspuru-Guzik, A.
\newblock 2018.
\newblock Automatic chemical design using a data-driven continuous
  representation of molecules.
\newblock {\em ACS central science} 4(2):268--276.

\bibitem[\protect\citeauthoryear{Graves \bgroup et al\mbox.\egroup
  }{2017}]{graves2017automated}
Graves, A.; Bellemare, M.~G.; Menick, J.; Munos, R.; and Kavukcuoglu, K.
\newblock 2017.
\newblock Automated curriculum learning for neural networks.
\newblock {\em arXiv preprint arXiv:1704.03003}.

\bibitem[\protect\citeauthoryear{Guimaraes \bgroup et al\mbox.\egroup
  }{2017}]{guimaraes2017objective}
Guimaraes, G.~L.; Sanchez-Lengeling, B.; Outeiral, C.; Farias, P. L.~C.; and
  Aspuru-Guzik, A.
\newblock 2017.
\newblock Objective-reinforced generative adversarial networks (organ) for
  sequence generation models.
\newblock {\em arXiv preprint arXiv:1705.10843}.

\bibitem[\protect\citeauthoryear{Kadurin \bgroup et al\mbox.\egroup
  }{2017}]{kadurin2017cornucopia}
Kadurin, A.; Aliper, A.; Kazennov, A.; Mamoshina, P.; Vanhaelen, Q.; Khrabrov,
  K.; and Zhavoronkov, A.
\newblock 2017.
\newblock The cornucopia of meaningful leads: Applying deep adversarial
  autoencoders for new molecule development in oncology.
\newblock {\em Oncotarget} 8(7):10883.

\bibitem[\protect\citeauthoryear{Kingma and Welling}{2013}]{kingma2013auto}
Kingma, D.~P., and Welling, M.
\newblock 2013.
\newblock Auto-encoding variational bayes.
\newblock {\em arXiv preprint arXiv:1312.6114}.

\bibitem[\protect\citeauthoryear{Kullback and
  Leibler}{1951}]{kullback1951information}
Kullback, S., and Leibler, R.~A.
\newblock 1951.
\newblock On information and sufficiency.
\newblock {\em The annals of mathematical statistics} 22(1):79--86.

\bibitem[\protect\citeauthoryear{Landrum and others}{2006}]{landrum2006rdkit}
Landrum, G., et~al.
\newblock 2006.
\newblock Rdkit: Open-source cheminformatics.

\bibitem[\protect\citeauthoryear{Lillicrap \bgroup et al\mbox.\egroup
  }{2015}]{lillicrap2015continuous}
Lillicrap, T.~P.; Hunt, J.~J.; Pritzel, A.; Heess, N.; Erez, T.; Tassa, Y.;
  Silver, D.; and Wierstra, D.
\newblock 2015.
\newblock Continuous control with deep reinforcement learning.
\newblock {\em arXiv preprint arXiv:1509.02971}.

\bibitem[\protect\citeauthoryear{Lim \bgroup et al\mbox.\egroup
  }{2018}]{lim2018molecular}
Lim, J.; Ryu, S.; Kim, J.~W.; and Kim, W.~Y.
\newblock 2018.
\newblock Molecular generative model based on conditional variational
  autoencoder for de novo molecular design.
\newblock {\em arXiv preprint arXiv:1806.05805}.

\bibitem[\protect\citeauthoryear{Mnih \bgroup et al\mbox.\egroup
  }{2015}]{mnih2015human}
Mnih, V.; Kavukcuoglu, K.; Silver, D.; Rusu, A.~A.; Veness, J.; Bellemare,
  M.~G.; Graves, A.; Riedmiller, M.; Fidjeland, A.~K.; Ostrovski, G.; et~al.
\newblock 2015.
\newblock Human-level control through deep reinforcement learning.
\newblock {\em Nature} 518(7540):529.

\bibitem[\protect\citeauthoryear{Narvekar and
  Stone}{2019}]{narvekar2019learning}
Narvekar, S., and Stone, P.
\newblock 2019.
\newblock Learning curriculum policies for reinforcement learning.
\newblock In {\em Proceedings of the 18th International Conference on
  Autonomous Agents and MultiAgent Systems},  25--33.
\newblock International Foundation for Autonomous Agents and Multiagent
  Systems.

\bibitem[\protect\citeauthoryear{Narvekar}{2017}]{narvekar2017curriculum}
Narvekar, S.
\newblock 2017.
\newblock Curriculum learning in reinforcement learning.
\newblock In {\em IJCAI},  5195--5196.

\bibitem[\protect\citeauthoryear{Neil \bgroup et al\mbox.\egroup
  }{2018}]{neil2018exploring}
Neil, D.; Segler, M.; Guasch, L.; Ahmed, M.; Plumbley, D.; Sellwood, M.; and
  Brown, N.
\newblock 2018.
\newblock Exploring deep recurrent models with reinforcement learning for
  molecule design.

\bibitem[\protect\citeauthoryear{Olivecrona \bgroup et al\mbox.\egroup
  }{2017}]{olivecrona2017molecular}
Olivecrona, M.; Blaschke, T.; Engkvist, O.; and Chen, H.
\newblock 2017.
\newblock Molecular de-novo design through deep reinforcement learning.
\newblock {\em Journal of cheminformatics} 9(1):48.

\bibitem[\protect\citeauthoryear{Popova, Isayev, and
  Tropsha}{2018}]{popova2018deep}
Popova, M.; Isayev, O.; and Tropsha, A.
\newblock 2018.
\newblock Deep reinforcement learning for de novo drug design.
\newblock {\em Science advances} 4(7):eaap7885.

\bibitem[\protect\citeauthoryear{Segler \bgroup et al\mbox.\egroup
  }{2017}]{segler2017generating}
Segler, M.~H.; Kogej, T.; Tyrchan, C.; and Waller, M.~P.
\newblock 2017.
\newblock Generating focused molecule libraries for drug discovery with
  recurrent neural networks.
\newblock {\em ACS central science} 4(1):120--131.

\bibitem[\protect\citeauthoryear{Tanimoto}{1958}]{tanimoto1958elementary}
Tanimoto, T.~T.
\newblock 1958.
\newblock Elementary mathematical theory of classification and prediction.

\bibitem[\protect\citeauthoryear{Weininger, Weininger, and
  Weininger}{1989}]{weininger1989smiles}
Weininger, D.; Weininger, A.; and Weininger, J.~L.
\newblock 1989.
\newblock Smiles. 2. algorithm for generation of unique smiles notation.
\newblock {\em Journal of Chemical Information and Computer Sciences}
  29(2):97--101.

\bibitem[\protect\citeauthoryear{Zhou \bgroup et al\mbox.\egroup
  }{2019}]{zhou2019optimization}
Zhou, Z.; Kearnes, S.; Li, L.; Zare, R.~N.; and Riley, P.
\newblock 2019.
\newblock Optimization of molecules via deep reinforcement learning.
\newblock {\em Scientific reports} 9(1):10752.

\bibitem[\protect\citeauthoryear{Zhou, Li, and Zare}{2017}]{zhou2017optimizing}
Zhou, Z.; Li, X.; and Zare, R.~N.
\newblock 2017.
\newblock Optimizing chemical reactions with deep reinforcement learning.
\newblock {\em ACS central science} 3(12):1337--1344.

\end{thebibliography}
\bibliographystyle{aaai}
\end{document}